\documentclass{article}

\usepackage{microtype}
\usepackage{graphicx}
\usepackage{subcaption}
\usepackage{booktabs} 

\usepackage{hyperref}

\usepackage[preprint]{icml2026}

\usepackage{amsmath}
\usepackage{amssymb}
\usepackage{mathtools}
\usepackage{amsthm}

\usepackage{times}
\usepackage{latexsym}
\usepackage{enumitem}
\usepackage{multirow}
\usepackage{array}
\usepackage{amsfonts}
\usepackage{subcaption}
\usepackage{makecell}
\usepackage{soul}
\usepackage{xcolor,colortbl}
\usepackage{xspace}
\usepackage{textcomp}
\usepackage{stfloats}
\usepackage{url}
\usepackage{verbatim}
\usepackage{graphicx}
\usepackage{algorithm}
\usepackage{algorithmic}
\usepackage{amssymb}
\usepackage{helvet}  
\usepackage{courier}  
\usepackage{newfloat}
\usepackage{listings}
\usepackage{multirow}
\usepackage{amsmath}
\usepackage{booktabs}
\usepackage{pifont}
\usepackage{color}
\usepackage{bbding}
\usepackage{listings}
\usepackage{fancyvrb}
\usepackage{pmboxdraw}
\usepackage{hyperref, cleveref}
\usepackage{xcolor}
\usepackage{tcolorbox}
\usepackage[export]{adjustbox}
\usepackage[table]{xcolor}
\definecolor{c1}{HTML}{cddef7}
\definecolor{c2}{HTML}{f5e1ea}
\definecolor{c3}{HTML}{efefef}

\definecolor{darkblue}{rgb}{0, 0, 0.5}
\hypersetup{colorlinks=true, citecolor=darkblue, linkcolor=darkblue, urlcolor=darkblue}

\theoremstyle{plain}

\theoremstyle{definition}

\theoremstyle{remark}

\usepackage[textsize=tiny]{todonotes}

\icmltitlerunning{\textsc{HeavySkill}: Heavy Thinking as the Inner Skill in Agentic Harness}

\begin{document}

\twocolumn[
  \icmltitle{\textsc{HeavySkill}: Heavy Thinking as the Inner Skill in Agentic Harness}




  \icmlsetsymbol{equal}{*}

  \begin{icmlauthorlist}
    \icmlauthor{Jianing Wang}{yyy}
    \icmlauthor{Linsen Guo}{yyy}
    \icmlauthor{Zhengyu Chen}{yyy}
    \icmlauthor{Qi Guo}{yyy,sch}
    \icmlauthor{Hongyu Zang}{yyy}
    \icmlauthor{Wenjie Shi}{yyy}
    \icmlauthor{Haoxiang Ma}{yyy}
    \icmlauthor{Xiangyu Xi}{yyy}
    \icmlauthor{Xiaoyu Li}{yyy}
    \icmlauthor{Wei Wang}{yyy}
    \icmlauthor{Xunliang Cai}{yyy}
  \end{icmlauthorlist}

  \icmlaffiliation{yyy}{Meituan LongCat Team}
  \icmlaffiliation{sch}{National Engineering Research Center for Software Engineering, Peking University, Beijing, China}

  \icmlcorrespondingauthor{Jianing Wang}{lygwjn@gmail.com}

  \icmlkeywords{Agentic Harness, Heavy Thinking, Large Language Model}

  \vskip 0.3in
]



\printAffiliationsAndNotice{}  

\begin{abstract}


Recent advances in agentic harness with orchestration frameworks that coordinate multiple agents with memory, skills, and tool use have achieved remarkable success in complex reasoning tasks. However, the underlying mechanism that truly drives performance remains obscured behind intricate system designs. 
In this paper, we propose \textsc{HeavySkill}~\footnote{\url{https://github.com/wjn1996/HeavySkill}.}, a perspective that views heavy thinking not only as a minimal execution unit in orchestration harness but also as an inner skill internalized within the model’s parameters that drives the orchestrator to solve complex tasks. 
We identify this skill as a two-stage pipeline, i.e., parallel reasoning then summarization, which can operate beneath any agentic harness. 
We present a systematic empirical study of \textsc{HeavySkill} across diverse domains. Our results show that this inner skill consistently outperforms traditional Best-of-N (BoN) strategies; notably, stronger LLMs can even approach Pass@N performance. 
Crucially, we demonstrate that the depth and width of heavy thinking, as a learnable skill, can be further scaled via reinforcement learning, offering a promising path toward self-evolving LLMs that internalize complex reasoning without relying on brittle orchestration layers.
\end{abstract}

\section{Introduction}

Recently, large language model (LLM) agents have demonstrated remarkable success on solving complex reasoning tasks via an orchestrated harness~\citep{Meng2025Agent, wang2024survey}, reinforcement learning with verified rewards (RLVR)~\citep{guo2025deepseek, Zheng2025Group}, and self-evolving learning~\citep{gao2026surveyselfevolvingagentswhat, wang2026longcatflashproveradvancingnativeformal, Wang2024Self}.
To better guide the LLM agent to perform task execution, Claude Code~\citep{TheC3} develops \textit{skills library} to inject extended knowledge and reusable strategies to the model with optional RLVR~\citep{xu2026agentskillslargelanguage}.
Inspired by this technique, multiple flexible harnesses have been proposed, such as CodeX~\citep{chen2021evaluatinglargelanguagemodels}, Claude Code~\citep{TheC3}, OpenClaw~\citep{openclaw}, and Hermes~\citep{hermes}.
Under this harness, the LLM acts as multiple different agents, and performs complex tasks through an orchestrator and accompanying \emph{}{Skill} and \emph{Memory} components.
However, the underlying mechanism that truly
drives performance remains obscured behind in-
tricate system designs.

Looking back at common harness models, orchestrator models typically operate within an agent loop, activating multiple subagents to execute various tasks in parallel based on user instructions and planning protocols, and ultimately summarizing the results. 
We believe this mode can be simplified into a two-stage workflow of \emph{parallel thinking} and \emph{summarization}.
In a word,
\textbf{we abstract the agentic harness into the LLM's inherent capability of heavy thinking}.
This approach revealed that the pursuit of empowering LLMs' reasoning capability focused on extensive parallel reasoning, a test-time scaling (TTS) strategy that amplifies computational resources during the inference phase.
This success underscores a fundamental insight: LLMs can substantially benefit from exploring multiple reasoning trajectories before converging to a final answer, mirroring the cognitive process of human collective deliberation.
Recent efforts on parallel reasoning have primarily relied on specialized architectural modifications, reasoning pattern design, and large-scale post-training recipes~\citep{Pan2025Learning, Liu2024APAR, Jin2025Learning, Rodionov2025Hogwild, Hsu2025Group, Yang2025Multiverse, Yang2025APE, Zheng2025Parallel, Wen2025ParaThinker}. 
Specifically, these methods~\citep{Hsu2025Group, Zheng2025Parallel, Wen2025ParaThinker} modify the existing thinking pattern with multiple inline thinking tags to elicit the LLMs to derive multiple trajectories simultaneously, following a summary stage to aggregate the different rationales to the final answer. 
In contrast, alternative frameworks such as Kimi K2~\citep{Bai2025Kimi}, and PaCoRe~\citep{pacore2025StepFun}, ane LongCat-Flash-Thinking-2601~\citep{wang2026longcatflashproveradvancingnativeformal} have demonstrated promising results by decomposing heavy thinking into two distinct stages: a parallel reasoning stage provides some independent reasoning trajectories, followed by a sequential deliberation stage that aggregates all trajectories and outputs a final answer.

In this paper, we conduct a systematic empirical investigation of heavy thinking skill for orchestrated harness, and propose \textsc{HeavySkill} to consolidate the insights into a readable skill for LLM.
We first provide a simple but effective training-free framework that decomposes the heavy thinking into two separate phases, i.e., parallel reasoning, and sequential deliberation. 
In this framework, we also introduce a memory cache mechanism to store and organize reasoning trajectories, enabling iterative deliberation where the model progressively refines its reasoning by revisiting and synthesizing prior attempts. 
Through extensive experiments spanning STEM, coding, and general tasks, we observe that heavy thinking substantially outperforms single reasoning.
On STEM-oriented benchmarks with verifiable numerical answers, we also show a consistent performance hierarchy: Heavy-Pass@k $\geq$ Heavy-Mean@K $\geq$ Vote@K $\geq$ Mean@k. 
Notably, models with stronger intrinsic reasoning abilities can approach Pass@$k$ upper bounds under heavy thinking, suggesting that sequential deliberation enables effective identification and synthesis of correct reasoning paths. 
Qualitative analysis reveals that models explicitly compare trajectory differences during deliberation, functioning as implicit verifiers.


Further analysis investigates which components substantially contribute towards the final performance. Ablation studies illuminate the complementary roles of framework components.
We find that the quality and diversity of trajectories generated from parallel reasoning stage are two keys for the performance.
We also show that sequential deliberation almost relies on the general capability of the model employed in this stage, suggesting that separate optimization of thinking and deliberation models may yield additional gains.
In addition, we demonstrate that reinforcement learning from verifiable rewards (RLVR) can be adapted to optimize both reasoning breadth (via parallel generation) and depth (via deliberation), simultaneously improving Heavy-Mean@$k$ and Pass@$k$ metrics. 

The main contributions are shown in the following.
This work makes three primary contributions: 
1) We introduce a simple but effective training-free framework for reproducing heavy thinking through parallel-reasoning and sequential deliberation.
2) We are the first to conduct the comprehensive empirical study to exhibit the performance of heavy thinking across diverse model scales and task domains, establishing its effectiveness and limitations.
3) We provide systematic analyses and insights into the interplay between framework components, and demonstrate the potential of heavy-mode-aware reinforcement learning as a superior optimization paradigm for reasoning-centric LLMs.

\begin{figure*}[t]
  \includegraphics[width=\linewidth]{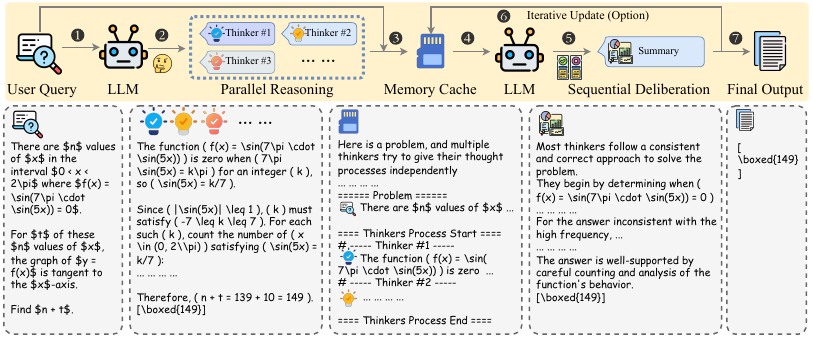} \hfill
  \caption{The overview framework of heavy thinking in LLMs test time scaling.}
  \label{fig:framework}
\end{figure*}

\section{Methodology}

\subsection{Workflow of Heavy Thinking}
\label{sec:workflow}

We thus describe the framework of heavy thinking. The overview of architecture is shown in Figure~\ref{fig:framework}.
The inference pipeline is decomposed into two separate phases, including parallel reasoning, and sequential deliberation.

Given a problem $q$, the goal of the parallel reasoning phase is to produce multiple independent trajectories. Formally, we can obtain $\mathcal{T}_{\pi_{\theta}}(q, K)=\{y_1, \cdots, y_{K}\}$, where $K$ denotes the number of trajectories, $\pi_{\theta}$ represents the LLM that aims to produce these trajectories, $y_i=\{\pi_{\theta}(y_{ij}|q, y_{i,<j})\}_{j=1}$ is one of the generated trajectories.

When the parallel reasoning is finished, we choose another LLM $\pi_{\phi}$ to produce a summary content in the sequential deliberation, which can be viewed as a second-time reasoning process that aggregates these trajectories derived from $\pi_{\theta}$.
Formally, we can obtain $\mathcal{T}_{\pi_{\phi}}(x_{c}, K^{(1)})$, where $x_c=\mathcal{C}(\mathcal{T}_{\pi_{\theta}}(q, K))$ denotes the serialized memory cache derived from parallel reasoning, $K^{(1)}$ represents the number of generated summary content.
We will describe this cache in the following section.

\subsection{Serialized Memory Cache}

To seamlessly bridge the two phases, we introduce a memory cache mechanism, which is a serialized context to store the candidate trajectories generated from the framework in history.
Since each trajectory generated by reasoning models typically encompasses both extensive internal thinking content and answer content, serializing all complete trajectories would exceed the model's maximum length limit. 
To ensure the robustness of subsequent inference, pruned trajectories are shuffled to prevent the model from developing a bias toward specific positions in the prompt.
To this end, we define the serialized context as $\mathcal{C}(x_c)$, establishing it as the input for the sequential deliberation stage. 
The specific prompting function is shown in Appendix~\ref{app:memory}.

\subsection{Iterative Deliberation}

We also introduce iterative deliberation, inspired by human behaviors in the real-world that repeatedly refine the ideas that were previously considered.
Specifically, at the $t\in\{2, \cdots, N\}$ iteration, we modify the memory cache by concatenating a loop input from the previous content from the sequential deliberation, i.e., $\mathcal{C}(x^{(t)})$, where $x_c^{(t)}=\mathcal{T}_{\pi_{\phi}}(x_c^{(t-1)}, K^{(t-1)}) || x^{(t-1)}$ is the modified cache, $K^{(t-1)}$ is the number of generated summary content phase, $\cdot||\cdot$ is the concatenation operation, $N$ is the total number of the iteration.

\subsection{Readable Skill for Agentic Harness}
\label{sec:readable_skill}

This workflow provides a concrete Python pipeline for executing heavy thinking. However, modern agentic harnesses---such as Claude Code~\citep{TheC3}, CodeX~\citep{chen2021evaluatinglargelanguagemodels}, and Hermes~\citep{hermes}---organize capabilities as \emph{skills}: human-readable, model-interpretable documents that the orchestrator loads into its context window at inference time. A skill specifies \emph{when} to activate, \emph{how} to execute, and \emph{what} to output, without requiring any code modification to the harness itself. This motivates us to distill the heavy thinking workflow into a single readable skill file.

\paragraph{Skill Structure.}
A readable skill is a structured natural-language document that serves as an executable specification for the LLM orchestrator. The \textsc{HeavySkill} document consists of four components:

\begin{itemize}[leftmargin=*,nosep]
\item \textbf{Activation Conditions} A declarative description of when heavy thinking should be triggered. The skill instructs the orchestrator to activate when facing tasks that involve complex reasoning and to remain dormant for simple factual queries or casual conversation. This conditional activation ensures that the additional inference cost is only incurred when the task complexity justifies it.

\item \textbf{Parallel Reasoning Protocol} Instructions for the orchestrator to spawn $K$ independent reasoning agents in parallel, each solving the same problem from scratch without access to other agents' outputs. The skill encourages diversity by suggesting that agents employ different problem-solving strategies (e.g., algebraic versus geometric approaches). In the harness context, each agent corresponds to a subagent call, which is natively supported by modern orchestration frameworks.

\item \textbf{Deliberation Prompt} The core of the skill is a carefully designed prompt template for the sequential deliberation stage. This prompt, which corresponds to the ``General-Prompt'' in our workflow implementation, instructs the deliberation model $\pi_\phi$ to: 1) \emph{classify the query type} to determine the appropriate analysis depth; 2) \emph{critically evaluate} each thinker's reasoning rather than naively following the majority; 3) \emph{re-derive} the answer when all thinkers are judged to be incorrect; and 4) \emph{maintain language and format consistency} with the original query. The prompt explicitly prohibits superficial concatenation of thinker outputs and instead demands genuine synthesis. The full prompt is presented in Figure~\ref{fig:heavy_prompt} (Appendix~\ref{app:memory}).

\item \textbf{Output Constraints} The skill specifies that the final response must contain only the answer---not the meta-analysis---and must follow the format conventions of the target domain (e.g., $\backslash boxed\{\cdot\}$ for mathematics, code blocks for programming).
\end{itemize}

\paragraph{From Workflow to Skill}
The key distinction between the workflow (Section~\ref{sec:workflow}) and the readable skill lies in the locus of control. In the workflow mode, an external Python pipeline orchestrates API calls, manages the memory cache, and routes outputs between stages. In the skill mode, the LLM orchestrator \emph{itself} reads the skill document and autonomously executes the prescribed protocol---spawning parallel agents, collecting their outputs into its context window as a serialized memory cache, and performing deliberation in a subsequent generation step. This self-orchestration is made possible by the in-context learning capability of frontier LLMs, which can faithfully follow multi-step procedural instructions embedded in their prompt.

\paragraph{Portability and Generality}
Because the readable skill is a plain-text document with no framework-specific dependencies, it can be injected into any harness that supports skill loading and subagent spawning. We have verified that the same \textsc{HeavySkill} document functions correctly under both Claude Code and custom orchestration harnesses, without modification. This portability aligns with our central thesis: heavy thinking is not an artifact of a particular system design but an \emph{inner skill} that can be activated across diverse orchestration environments. By encapsulating the two-stage pipeline as a transferable skill, we decouple the reasoning capability from the infrastructure, enabling any sufficiently capable LLM to perform heavy thinking.

\section{Experiments}

\begin{table*}[!t]
\footnotesize
\renewcommand{\arraystretch}{1.5}

\setlength{\tabcolsep}{0.8mm}
\resizebox{\linewidth}{!}{
\begin{tabular}{c | c | ccccc | ccccc | ccccc | ccccc }
\toprule
\multirow{2}{*}{\textbf{Models}}
&
\multirow{2}{*}{\textbf{$k$}}
& \multicolumn{5}{c|}{\textbf{AIME25}}
& \multicolumn{5}{c|}{\textbf{BeyondAIME}}
& \multicolumn{5}{c|}{\textbf{HMMT25-Feb}}
& \multicolumn{5}{c}{\textbf{GPQA-Diamond}} \\
\cmidrule(lr){3-7} \cmidrule(lr){8-12} \cmidrule(lr){13-17} \cmidrule(lr){18-22} 
&
& \makecell{M$@k$}
& \makecell{P$@k$}
& \makecell{V$@k$}
& \makecell{HM$@4$}
& \makecell{HP$@4$}
& \makecell{M$$@k$$}
& \makecell{P$@k$}
& \makecell{V$@k$}
& \makecell{HM$@4$}
& \makecell{HP$@4$}
& \makecell{M$@k$}
& \makecell{P$@k$}
& \makecell{V$@k$}
& \makecell{HM$@4$}
& \makecell{HP$@4$}
& \makecell{M$@k$}
& \makecell{P$@k$}
& \makecell{V$@k$}
& \makecell{HM$@4$}
& \makecell{HP$@4$}
\\
\midrule

\multicolumn{22}{c}{\bf \emph{Leading Close-Weights Models}} \\
\midrule

\multirow{2}{*}{\makecell{GPT-5 \\Thinking}}
& 8 & 
92.5 & 100 & 96.7 &\cellcolor{c1} 96.7 &\cellcolor{c1} 96.7 &
69.9 & 86.0 & 67.0 &\cellcolor{c1} 79.5 &\cellcolor{c1} 83.0 & 
90.4 & 96.7 & 93.3 &\cellcolor{c3} 93.1 &\cellcolor{c2} 96.7 & 
85.8 & 96.5 & 89.9 &\cellcolor{c1} 89.9 &\cellcolor{c1} 91.7\\

& 16 & 
91.9 & 100 & 96.7 &\cellcolor{c1} 99.2 &\cellcolor{c2} 100 &  
70.1 & 91.0 & 73.0 &\cellcolor{c1} 82.5 &\cellcolor{c1} 88.0 & 
89.8 & 96.7 & 86.7 &\cellcolor{c1} 95.0 &\cellcolor{c2} 96.7 &  
85.6 & 97.5 & 86.4 &\cellcolor{c1} 87.2 &\cellcolor{c1} 90.9 \\

\midrule
\multirow{2}{*}{\makecell{Claude 4.5 \\Thinking}}
& 8 & 
82.5 & 90.0 & 90.0 &\cellcolor{c2} 90.0 &\cellcolor{c2} 90.0 &  
58.9 & 74.0 & 66.0 &\cellcolor{c3} 63.3 &\cellcolor{c1} 67.5 &  
66.3 & 76.7 & 66.7 &\cellcolor{c1} 75.8 &\cellcolor{c2} 76.7 &  
82.1 & 91.9 & 80.3 &\cellcolor{c1} 83.1 &\cellcolor{c1} 86.9 \\

& 16 & 
83.1 & 93.3 & 90.0 &\cellcolor{c1} 90.0 &\cellcolor{c1} 90.0 &  
59.4 & 83.4 & 70.0 &\cellcolor{c3} 68.3 &\cellcolor{c1} 71.5 & 
69.4 & 86.7 & 80.0 &\cellcolor{c1} 85.8 &\cellcolor{c2} 86.7 &  
81.5 & 94.9 & 78.8 &\cellcolor{c1} 81.7 &\cellcolor{c1} 85.9 \\

\midrule
\multirow{2}{*}{\makecell{Gemini-3 Pro \\Preview}}
& 8 & 
95.0 & 96.7 & 96.7 &\cellcolor{c3} 95.8 &\cellcolor{c2} 96.7 & 
83.1 & 96.0 & 83.0 &\cellcolor{c1} 92.0 &\cellcolor{c1} 95.0 & 
95.4 & 100 & 100 &\cellcolor{c2} 100 &\cellcolor{c2} 100 & 
- & - & - & - & -  \\

& 16 &  
94.8 & 96.7 & 96.7 &\cellcolor{c3} 95.8 &\cellcolor{c2} 96.7 & 
- & - & - & - & -  & 
96.5 & 100  & 93.3 &\cellcolor{c2} 100  &\cellcolor{c2}  100 & 
- & - & - & - & -  \\

\midrule
\multicolumn{22}{c}{\bf \emph{Open-Weights Models}} \\
\midrule
\multirow{2}{*}{\makecell{R1-Distill-\\Qwen-7B}}
& 8 & 
42.1 & 66.7 & 50.0 &\cellcolor{c1} 50.0 &\cellcolor{c1} 56.7 &  
26.5 & 54.0 & 32.0 &\cellcolor{c3} 30.8 &\cellcolor{c1} 39.0 &  
32.1 & 56.7 & 43.3 &\cellcolor{c3} 32.3 &\cellcolor{c3} 37.3 &  
48.8 & 85.9 & 49.5  &\cellcolor{c1} 51.5 &\cellcolor{c1} 63.1 \\

& 16 & 
41.7 & 66.7 & 60.0 &\cellcolor{c3} 56.7 &\cellcolor{c1} 60.0 &  
28.1 & 59.0 & 36.0 &\cellcolor{c3} 35.3 &\cellcolor{c1} 45.0 &  
29.2 & 80.0 & 40.0 &\cellcolor{c3} 31.7 &\cellcolor{c1} 43.3 &  
49.0 & 89.9 & 51.1 &\cellcolor{c1} 51.8 &\cellcolor{c1} 65.7 \\

\midrule
\multirow{2}{*}{\makecell{R1-Distill-\\Qwen-32B}}
& 8 & 
53.3 & 76.7 & 63.3 &\cellcolor{c1} 63.3 &\cellcolor{c1} 66.7 &  
30.3 & 57.0 & 40.0 &\cellcolor{c1} 41.0 &\cellcolor{c1} 47.0 &  
21.3 & 46.7 & 33.3 &\cellcolor{c1} 39.2 &\cellcolor{c1} 43.3 &  
64.5 & 88.4 & 66.7 &\cellcolor{c1} 66.9 &\cellcolor{c1} 71.7 \\

& 16 &  
52.3 & 80.0 & 66.7 &\cellcolor{c1} 68.3 &\cellcolor{c1} 76.7 &  
31.4 & 59.0 & 46.0 &\cellcolor{c3} 44.3 &\cellcolor{c1} 49.0 &  
23.3 & 63.3 & 43.3 &\cellcolor{c1} 45.8 &\cellcolor{c1} 50.0 &  
64.3 & 88.4 & 66.7 &\cellcolor{c1} 67.2 &\cellcolor{c1} 71.7 \\

\midrule
\multirow{2}{*}{\makecell{R1-Distill-\\Qwen3-8B}}
& 8 & 
76.7 & 90.0 & 83.3 &\cellcolor{c1} 85.8 &\cellcolor{c2} 90.0 &  
54.1 & 70.0 & 60.0 &\cellcolor{c3} 59.0 &\cellcolor{c1} 65.0 &  
58.3 & 80.0 & 66.7 &\cellcolor{c3} 65.0 &\cellcolor{c1} 66.7 &  
61.2 & 85.4 & 62.1 &\cellcolor{c1} 64.8 &\cellcolor{c1} 72.2 \\

& 16 & 
73.3 & 	86.6 & 	80.0 &\cellcolor{c1} 80.8 &\cellcolor{c1} 83.3 &  
52.8 & 	76.0 & 	58.0 &\cellcolor{c3} 56.5 &\cellcolor{c1} 61.0 &  
57.9 & 	86.6 & 	60.0 &\cellcolor{c1} 68.3 &\cellcolor{c1} 73.3 &  
61.5 & 90.4 & 66.7 &\cellcolor{c1} 68.1 &\cellcolor{c1} 74.2 \\

\midrule
\multirow{2}{*}{\makecell{Qwen3-8B}}
& 8 & 
69.6 & 86.7 & 76.7 &\cellcolor{c1} 80.0 &\cellcolor{c1} 80.0 &  
46.1 & 66.0 & 53.0 &\cellcolor{c3} 52.5 &\cellcolor{c1} 58.0 &  
47.9 & 76.7 & 56.7 &\cellcolor{c1} 56.7 &\cellcolor{c1} 56.7 &  
59.2 & 80.3 & 62.1 &\cellcolor{c1} 63.3 &\cellcolor{c1} 66.2  \\

& 16 & 
70.0 & 86.7 & 80.0 &\cellcolor{c1} 80.8 &\cellcolor{c1} 83.3 &  
45.6 & 72.0 & 53.0 &\cellcolor{c3} 52.3 &\cellcolor{c1} 56.0 &  
49.0 & 83.3 & 56.7 &\cellcolor{c1} 58.3 &\cellcolor{c1} 60.0 &  
- & - & - & - & -    \\

\midrule
\multirow{2}{*}{\makecell{Qwen3-32B}}
& 8 & 
72.5 & 83.3 & 83.3 &\cellcolor{c3} 80.8 &\cellcolor{c2} 83.3 &  
52.0 & 70.0 & 58.0 &\cellcolor{c1} 58.5 &\cellcolor{c1} 62.0 &  
58.8 & 83.3 & 73.3 &\cellcolor{c3} 60.8 &\cellcolor{c1} 70.0 &  
69.8 & 85.4 & 70.7 &\cellcolor{c1} 71.7 &\cellcolor{c1} 76.3 \\

& 16 & 
73.1 & 90.0 & 80.0 &\cellcolor{c1} 86.7 &\cellcolor{c1} 86.7 &  
51.9 & 77.0 & 58.0 &\cellcolor{c1} 58.5 &\cellcolor{c1} 65.0 &  
57.9 & 90.0 & 60.0 &\cellcolor{c1} 62.5 &\cellcolor{c1} 70.0 &  
69.0 & 88.4 & 69.7 &\cellcolor{c1} 70.3 &\cellcolor{c1} 76.3 \\

\midrule
\multirow{2}{*}{\makecell{DeepSeek\\R1-0528}}
& 8 & 
87.1 & 96.7 & 90.0 &\cellcolor{c1} 93.3 &\cellcolor{c1} 93.3 &  
67.3 & 84.0 & 73.0 &\cellcolor{c1} 74.0 &\cellcolor{c1} 77.0 &  
80.8 & 93.3 & 86.7 &\cellcolor{c1} 91.7 &\cellcolor{c2} 93.3 &  
80.6 & 93.4 & 82.3 &\cellcolor{c1} 84.6 &\cellcolor{c1} 87.9 \\

& 16 & 
87.3 & 96.7 & 90 &\cellcolor{c2} 96.7 &\cellcolor{c2} 96.7 &  
- & - & - & - & -  &
78.5 & 93.3 & 83.3 &\cellcolor{c1} 91.7 &\cellcolor{c2} 93.3 &  
80.2 & 94.4 & 81.3 &\cellcolor{c1} 83.3 &\cellcolor{c1} 85.9 \\

\midrule
\multirow{2}{*}{\makecell{GPT-\\OSS-20B}}
& 8 & 
92.2 & 96.7 & 96.7 &\cellcolor{c3} 92.5 &\cellcolor{c2} 96.7 &  
65.0 & 82.0 & 69.0 &\cellcolor{c3} 67.5 &\cellcolor{c1} 74.0 &  
78.3 & 90.0 & 90.0 &\cellcolor{c3} 83.3 &\cellcolor{c2} 93.3 &  
- & - & - & - & -    \\

& 16 & 
92.0 & 96.7 & 96.7 &\cellcolor{c3} 93.3 &\cellcolor{c2} 96.7 &  
- & - & - & - & -   & 
76.1 & 93.3 & 90.0 &\cellcolor{c3} 85.0 &\cellcolor{c1} 90.0 &  
- & - & - & - & -   \\

\midrule
\multirow{2}{*}{\makecell{Kimi K2\\Thinking}}
& 8 & 
95.4 & 100 & 96.7 &\cellcolor{c2} 100 &\cellcolor{c2} 100 & 
76.8 & 87.0 & 81.0 &\cellcolor{c1} 83.0 &\cellcolor{c1} 84.0  & 
87.5 & 100 & 90.0 &\cellcolor{c1} 93.3 &\cellcolor{c1} 93.3  & 
85.2 & 94.9 & 82.3 &\cellcolor{c1} 86.9 &\cellcolor{c1} 89.9 \\

& 16 & 
95.2 & 100 & 96.7 &\cellcolor{c1} 99.2 &\cellcolor{c2} 100 & 
- & - & - & - & -   & 
88.5 & 100 & 93.3 &\cellcolor{c3} 90.0 &\cellcolor{c1} 93.3 & 
85.3 & 97.0 & 80.3 &\cellcolor{c1} 87.5 &\cellcolor{c1} 91.4 \\

\midrule
\multirow{2}{*}{\makecell{GLM 4.6}}
& 8 &  
91.3 & 96.7 & 96.7 &\cellcolor{c2} 96.7 &\cellcolor{c2} 96.7 &  
74.1 & 90.0 & 78.0 &\cellcolor{c1} 78.8 &\cellcolor{c1} 81.0 & 
91.3 & 100 & 96.7 &\cellcolor{c2} 100 &\cellcolor{c2} 100 &  
82.9 & 93.9 & 79.8 &\cellcolor{c1} 85.2 &\cellcolor{c1} 88.9 \\

& 16 & 
93.1 & 100 & 100 &\cellcolor{c3} 96.7 &\cellcolor{c3} 96.7 &  
- & - & - & - & -   & 
90.4 & 100 & 96.7 &\cellcolor{c1} 99.2 &\cellcolor{c2} 100 &  
82.5 & 96.0 & 80.3 &\cellcolor{c1} 87.5 &\cellcolor{c1} 91.4 \\

\midrule
\multirow{2}{*}{\makecell{DeepSeek V3.2\\Thinking}}
& 8 & 
93.3 & 100 & 93.3 &\cellcolor{c1} 93.3 &\cellcolor{c1} 96.7 &  
- & - & - & - & -    & 
91.3 & 100 & 96.7 &\cellcolor{c1} 96.7 &\cellcolor{c1} 96.7 &  
- & - & - & - & -   \\

& 16 &  
93.5 & 100 & 96.7 &\cellcolor{c2} 100 &\cellcolor{c2} 100 &  
- & - & - & - & -   & 
92.3 & 100 & 93.3 &\cellcolor{c2} 100 &\cellcolor{c2} 100 &  
- & - & - & - & -   \\

\bottomrule
\end{tabular}
}
\centering
\caption{Overview performance of heavy mode on STEM tasks (Heavy Mean@4, simplify as HM@4) compared to basic TTS metrics in the parallel reasoning phase, i.e., Mean@K (M@K), Pass@K (P@K), and Vote@K (V@K).}
\label{tab:heavy-stem} 
\end{table*}

\subsection{Setups}

By default, both phases of the LLM use the same model (i.e., $\pi_{\theta}=\pi_{\phi}$) unless otherwise specified.
We choose multiple close-weight and open-weight models for evaluation. 
Concretely, the close-weights models consist of GPT-5-Thinking~\citep{gpt52025OpenAI}, Claude 4.5 Thinking, and Gemini 3 Pro Preview. The open-weights models contain R1-Distill-Qwen-7B~\citep{guo2025deepseek}, R1-Distill-Qwen-32B~\citep{guo2025deepseek}, R1-Distill-Qwen3-8B~\citep{guo2025deepseek}, Qwen3-8B~\citep{Yang2025Qwen3}, Qwen3-32B~\citep{Yang2025Qwen3}, DeepSeek R1-0528~\citep{guo2025deepseek}, GPT-OSS-20B~\citep{OpenAI2025gpt}, Kimi K2 Thinking~\citep{Bai2025Kimi}, GLM4.6~\citep{Zeng2025glm}, and DeepSeek V3.2 Thinking~\citep{deepseekai2025deepseekv32pushingfrontieropen}.

In the main experiments, we set temperature as 1.0, topp as 0.95 and topk=10. The number of iterations is $N=1$, the number of parallel trajectories $K\in\{8, 16\}$, and the number of generated summary content is $K^{(1)}=4$.
For the metrics, we choose three basic values: 1) Mean@K (M@K) denotes the average accuracy of the selected $K$ parallel trajectories from the parallel reasoning phase; 2) Pass@K (P@K) represents the proportion of the $K$ selected trajectories where at least one is correct, which can be used to measures the boundary of the model's inference ability. 3) Vote@K (V@K) denotes the accuracy of the trajectories with the highest frequency of answers, which is similar to BoN.
We also design two metrics: 1) Heavy-Mean@K (HM@K) denotes the average accuracy of the content after the second phase; 2) Heavy-Pass@K represents the proportion of the generated summary contents where at least one is correct.

\begin{table*}[!t]
\footnotesize
\renewcommand{\arraystretch}{1.5}

\setlength{\tabcolsep}{0.8mm}
\resizebox{\linewidth}{!}{
\begin{tabular}{c | cccc | cccc | cccc | cccc }
\toprule
\multirow{2}{*}{\textbf{Models}}
& \multicolumn{4}{c|}{\textbf{LiveCodeBench (24.08-25.05)}} 
& \multicolumn{4}{c|}{\textbf{Arena-Hard}}
& \multicolumn{4}{c|}{\textbf{IFEval}} 
& \multicolumn{4}{c}{\textbf{IMO (Answer Bench)}} \\
\cmidrule(lr){2-5} \cmidrule(lr){6-9} \cmidrule(lr){10-13} \cmidrule(lr){14-17}
& \makecell{M$@k$}
& \makecell{P$@k$}
& \makecell{HM$@4$}
& \makecell{HP$@4$}
& \makecell{M$@k$}
& \makecell{P$@k$}
& \makecell{HM$@4$}
& \makecell{HP$@4$}
& \makecell{M$@k$}
& \makecell{P$@k$}
& \makecell{HM$@4$}
& \makecell{HP$@4$}
& \makecell{M$@k$}
& \makecell{P$@k$}
& \makecell{HM$@4$}
& \makecell{HP$@4$}
\\
\midrule

\makecell{Qwen3-8B} &
55.5 & 67.4 & 56.8 & 63.8 &
 26.0 & - & 25.0 & - &
 85.4 & - & 85.2 & 92.4 &
 50.2 & 68.8 & 50.3 & 63.3 \\

\midrule

\makecell{R1-Distill-\\Qwen3-8B} &
56.3 & 70.9 & 56.8 & 67.4 &
20.8 & - & 18.7 & - &
35.7 & - & 69.3 & 86.5 &
47.0 & 72.5 & 47.2 & 62.3 \\



 \midrule

\makecell{GLM 4.6} &
81.0 & 90.3 & 81.3 & 87.9 &
88.2 & - & 88.1 & - &
88.8 & - & 88.5 & 94.8 &
74.5 & 89.5 & 75.1 & 86.0\\
 
 \midrule

\makecell{Kimi K2\\Thinking} &
81.2 & 91.0 & 83.7 & 80.4 &
83.5 & - & 83.1 & - &
92.5 & - & 92.0 & 97.6 &
69.1 & 85.3 & 77.2 & 88.0 \\
 
 \midrule

 \makecell{GPT-\\OSS-20B} &
69.7 & 89.0 & 69.2 & 85.5 &
25.4  & - & 25.0 & - &
90.8 & - & 91.1 & 97.6 &
65.8 & 81.5 & 71.0 & 84.5 \\
 
\bottomrule
\end{tabular}
}
\centering
\caption{Overview performance of heavy thinking on general reasoning tasks (Heavy Mean@4, simplify as HM@4) compared to basic TTS metrics in the parallel reasoning phase, i.e., Mean@K (M@K) and Pass@K (P@K).}
\label{tab:heavy-general} 
\end{table*}

\subsection{Evaluations on STEM Tasks}

In this section, we evaluate the effectiveness of the "heavy thinking" framework across a wide range of STEM tasks, including AIME25, BeyondAIME, HMMT25-Feb, and GPQA-Diamond.
We compare HM@4 and HP@4 metrics against standard test-time scaling metrics, such as such as the mean performance (M@K), the intrinsic potential (P@K), and Majority Voting (V@K).
The main results are shown in Table~\ref{tab:heavy-stem}.

\paragraph{Heavy thinking consistently outperforms single-trajectory attempts}
Our empirical results demonstrate that HM@4 consistently surpasses 
M@K across all models and STEM benchmarks.
This indicates that parallel reasoning combined with sequential deliberation invariably yields a positive performance gain over the average quality of individual reasoning trajectories. Notably, when employing large-scale frontier models (e.g., Kimi K2 Thinking, GPT-5-Thinking), the heavy thinking often facilitates near-perfect scores on several benchmarks. These results are consistent with recent technical reports suggesting that scaling test-time compute through deliberation is a robust path toward saturation on difficult reasoning tasks.

\paragraph{Validation of Test-Time Scaling Laws}
By scaling the number of parallel trajectories ($K$) and employing sequential deliberation, we observe that the model's performance does not merely plateau but continues to improve, effectively leveraging the increased inference budget. 
This confirms that heavy thinking serves as a practical realization of Test-Time Scaling, where the "width" of reasoning (parallel exploration) and the "depth" of deliberation (sequential synthesis) act as multipliers for the base model's capability. 
This scaling property is particularly crucial for complex reasoning tasks where a single inference pass is often insufficient, validating that allocating more compute at test time is a reliable strategy for boosting performance without retraining.

\paragraph{Superiority over heuristic voting strategies}
As highlighted by the blue cells in Table~\ref{tab:heavy-stem}, the performance of heavy thinking frequently exceeds that of the heuristic Majority Voting (V@K) strategy. This suggests that sequential deliberation is more effective at synthesizing and distilling the results of parallel reasoning paths than simple statistical consensus.
Interestingly, we observed that while highly capable models (e.g., DeepSeek R1-0528 and GLM-4.6) sometimes show parity with or slight underperformance compared to voting on AIME25, this is primarily due to a ceiling effect—these models already achieve exceptional scores (above 90), leaving minimal room for further differentiation. However, on more cognitively demanding benchmarks such as BeyondAIME, HMMT, and GPQA-Diamond, the advantage of the heavy thinking over voting becomes significantly more pronounced, underscoring its utility for complex problem-solving.

\paragraph{Potential to surpass intrinsic reasoning boundaries}

While it remains challenging for the aggregate performance (HM@4) to surpass the theoretical potential of the raw trajectories (P@K), our results show that HM@4 frequently approaches P@K in frontier models like DeepSeek V3.2 and GPT-5 Thinking.
Remarkably, with a sufficiently LLM in the deliberation process, the potential of the heavy thinking (HP@4) exceeds the raw thinking potential (P@K) in nearly half of our experimental trials. 
This suggests that the deliberation process does not merely select from existing answers but can synthesize cross-trajectory insights to generate correct solutions that were not present in any single raw reasoning path. 
This finding provides a strong empirical foundation for leveraging RLVR to further bridge the gap between HM@4 and HP@4, potentially pushing the limits of LLM reasoning beyond their inherent per-trajectory constraints.

\begin{figure}[t]
\begin{center}
  \includegraphics[width=.95\linewidth]{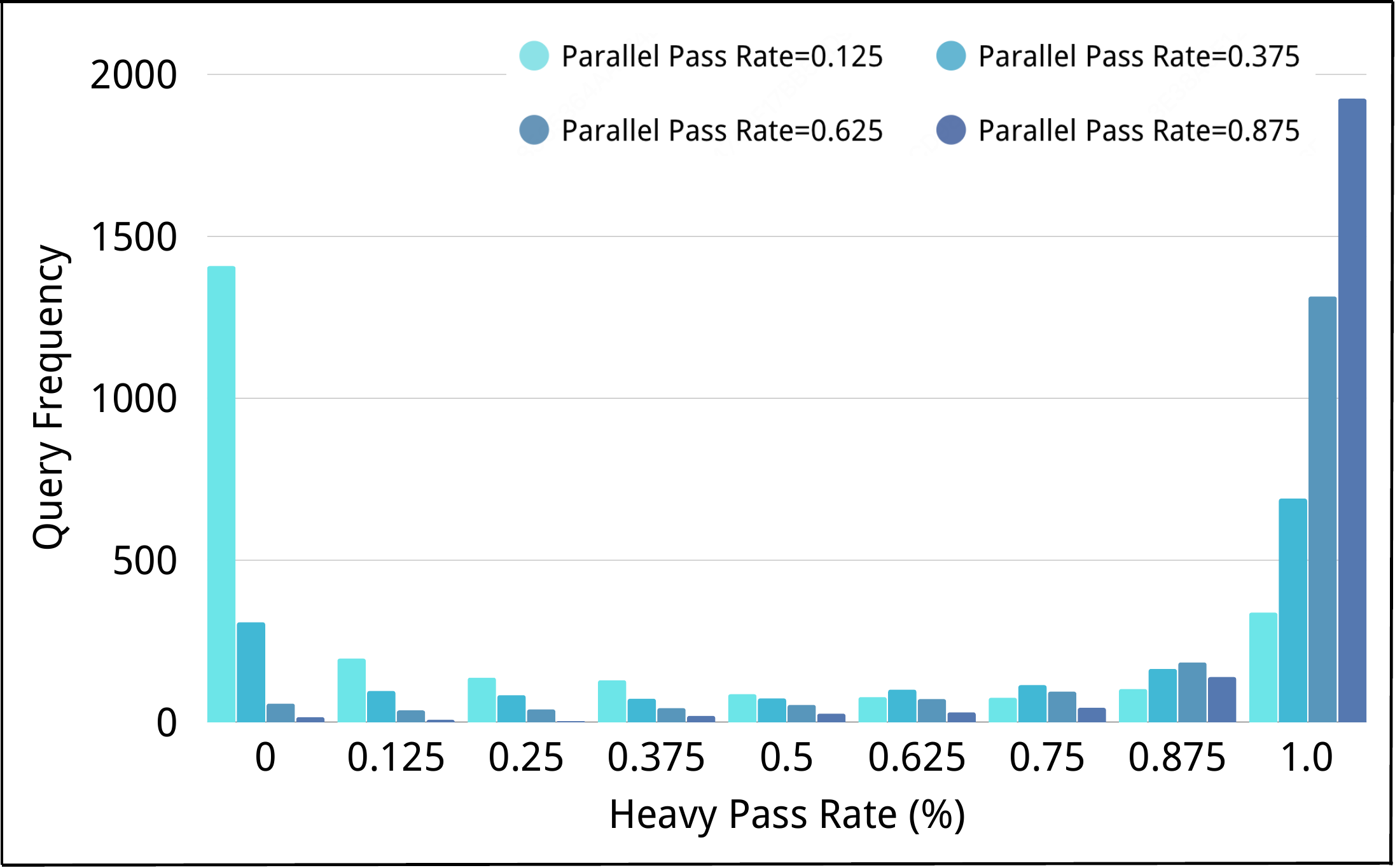} \hfill
  \caption{The pass rate distribution of heavy thinking in different pass rates of parallel reasoning.}
  \label{fig:heavy_pass_rate}
\end{center}
\end{figure}

\subsection{Evaluations on General Reasoning Tasks}


\paragraph{Task-Dependent Efficacy of Sequential Deliberation} Unlike the consistent gains observed in STEM tasks, the impact of the summary mechanism (HM@4) varies across general reasoning categories. On objective, verifiable tasks such as LiveCodeBench and IFEval, heavy thinking demonstrates substantial improvements. For instance, GPT-OSS-20B sees its performance surge from an M@K of 
69.7\% to an HM@4 of 85.5\% on LiveCodeBench. Similarly, R1-Distill-Qwen-32B experiences a significant boost on IFEval (35.7\% → 69.3\%). This confirms that for tasks with clear logical or programmatic constraints, the summary model effectively distills high-quality solutions from multiple reasoning paths.

\paragraph{Challenges in Subjective Alignment} 
On Arena-Hard, which focuses on human-like chat and open-ended preferences, the gains from 
HM@4 are more marginal or occasionally slightly negative. This suggests that while sequential deliberation excels at "correctness-oriented" tasks, its benefit is less pronounced in "preference-oriented" tasks where the "mean" of multiple responses may not necessarily align with the specific stylistic nuances favored by the reward model or judge.

\paragraph{Superiority of the Summary Potential} A key finding in Table~\ref{tab:heavy-general} is that the potential of the summary model (HP@4) consistently remains the highest metric across nearly all benchmarks. Notably, in tasks like IMO (Answer Bench), several models achieve HP@4>P@K (e.g., GLM 4.6 reaching 
86.0\% vs. 75.1\%). This indicates that the deliberation process is not merely selecting a winner from existing paths but has the capacity to "re-reason" and uncover correct answers that were initially missed in the raw P@K sampling.

\begin{figure*}
  \includegraphics[width=.8\linewidth]{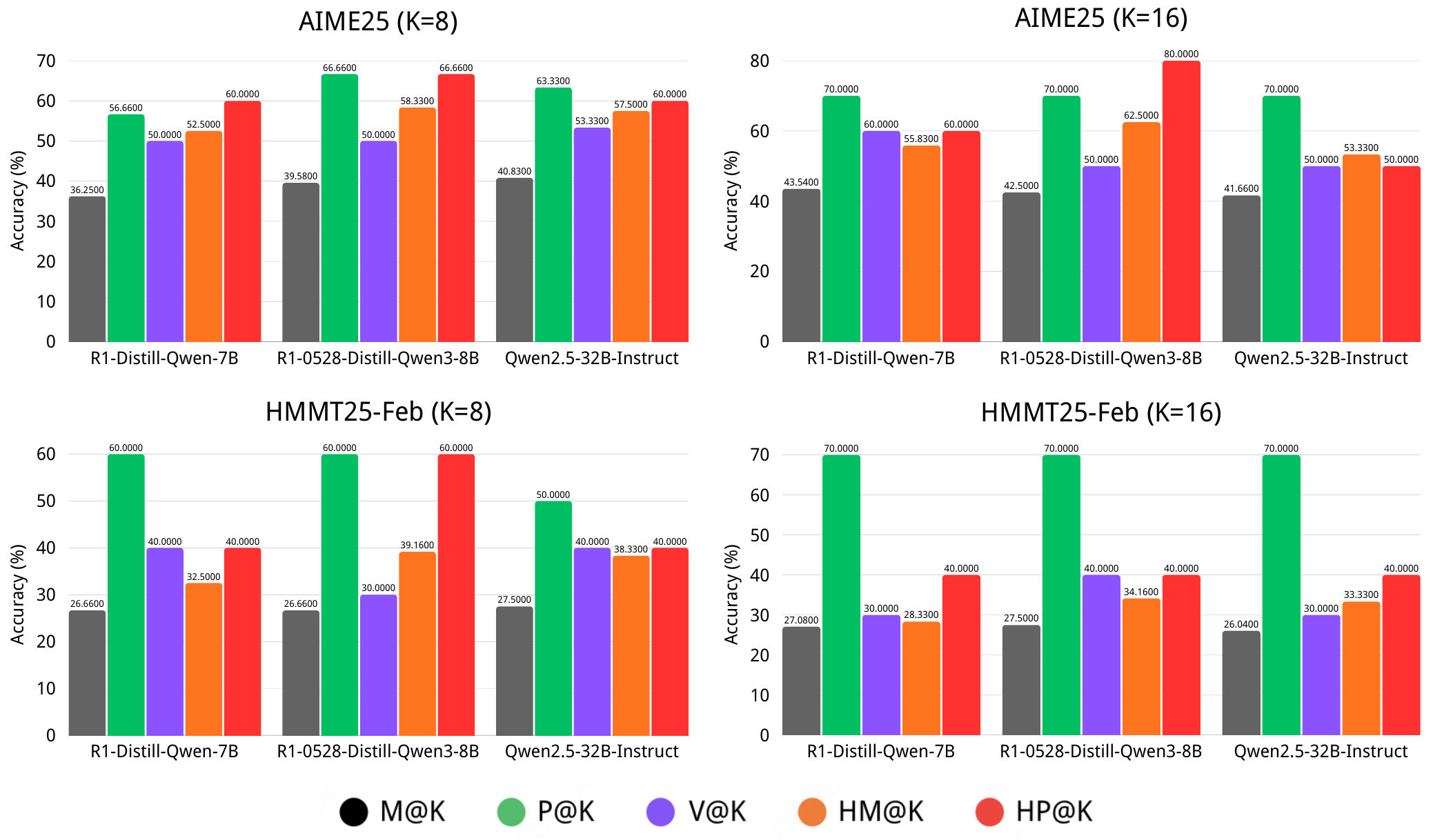} \hfill
  \centering
  \caption{When fixing the LLM as R1-Distill-Qwen-7B in the parallel reasoning phase, the final performance of different LLMs in sequential deliberation.}
  \label{fig:heavy_role}
\end{figure*}

\begin{figure*}[t]
  \includegraphics[width=.9\linewidth]{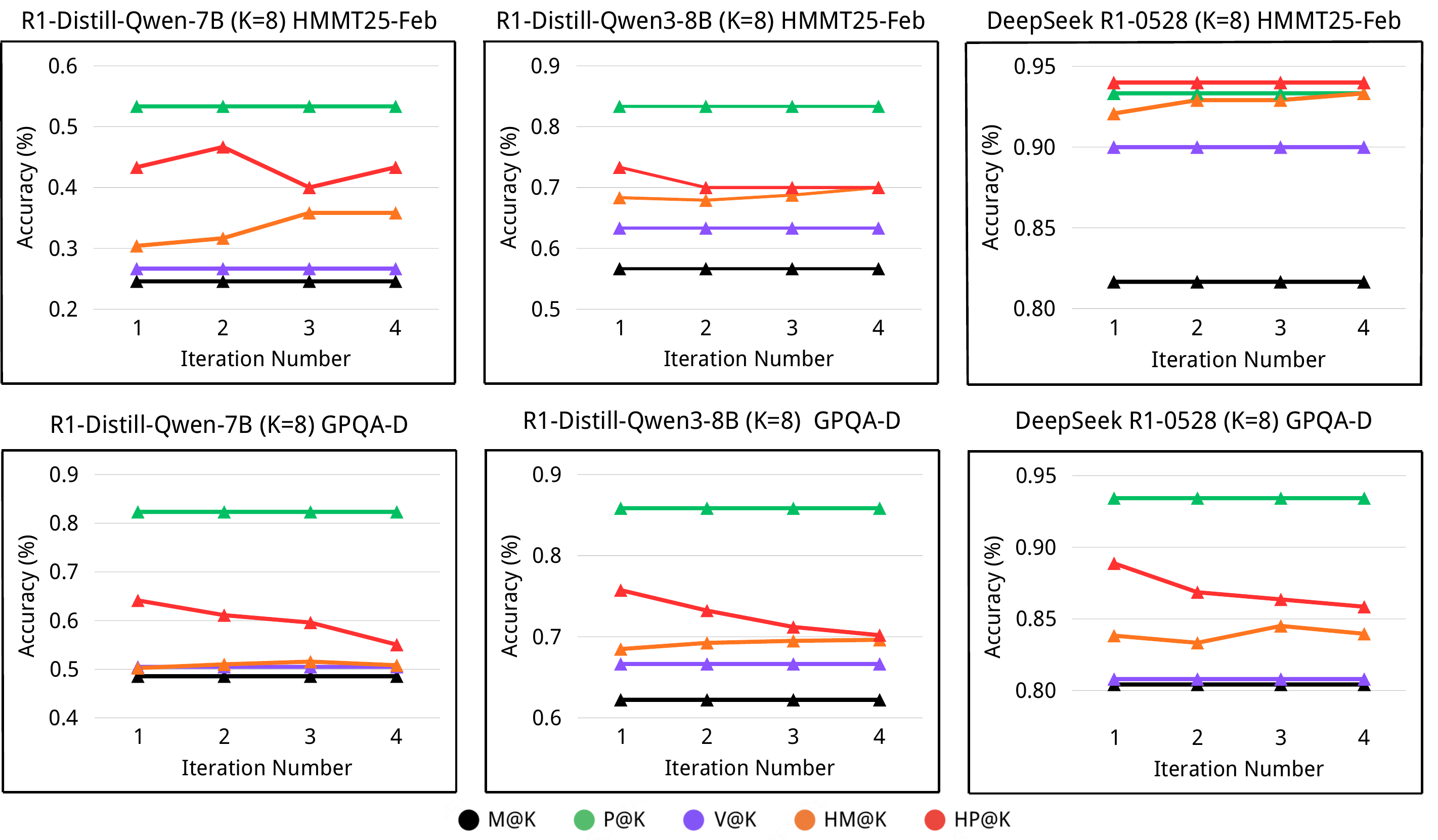} \hfill
  \centering
  \caption{The effectiveness of different numbers of iterations.}
  \label{fig:iterative_heavy}
\end{figure*}

\section{Further Analysis}

\subsection{Can Sequential Deliberation Revises Parallel Thinking?}

Our preliminary observations in Table~\ref{tab:heavy-stem} indicate that heavy thinking consistently outperforms vanilla majority voting, suggesting that the model possesses the intrinsic capability to discern and select correct answers even when they appear as low-frequency trajectories in parallel sampling. 
To further investigate this capability, we provide a granular analysis of the distributional relationship between the pass rates of parallel reasoning and heavy thinking.

Specifically, we choose open-resource data from Skywork OR1, DAPO, and DeepScaler, and leverage R1-Distill-Qwen-7B model as our experimental backbone.
We randomly sample 10k queries and conduct parallel reasoning with a sampling size of $K=16$ for each query to determine its baseline parallel pass rate. We then categorize queries into distinct groups based on specific parallel pass rate intervals $\{0.125,0.375,0.625,0.875\}$. 
For each group, we construct a corresponding memory cache and perform sequential deliberation (without iteration), the number of generated responses $K^{(1)}$ is 16.

Results in Figure~\ref{fig:heavy_pass_rate} illustrate the distribution of the heavy pass rate across different parallel pass rate cohorts. 
Our analysis yields the following key insights:
1) For queries with a parallel pass rate below 0.5, which typically pose a significant challenge for heuristic voting methods, heavy thinking demonstrates substantial corrective potential. 
Although approximately 1,400 queries remain unresolved, over 500 queries are successfully rectified through deliberation. 
This empirical evidence underscores the model's ability to refine errors even when the initial success probability is low.
2) In scenarios where the parallel pass rate exceeds 0.5, it maintains high accuracy, with a success rate exceeding 
98\%. 
While several queries (approximately 30) experience performance degradation, this loss is negligible compared to the overall performance gains achieved across the dataset.

\subsection{Model Selection for the Role of Summary}

To evaluate the robustness and generalizability of the framework, we investigate the performance when different models are paired for the two-stage reasoning process. 
Specifically, we fix the model for the parallel reasoning phase as R1-Distill-Qwen-7B. For the sequential deliberation stage, we select three models with varying architectures and scales: R1-Distill-Qwen-7B, R1-Distill-Qwen3-8B, and Qwen2.5-32B-Instruct. 

The experimental results are illustrated in Figure~\ref{fig:heavy_role}. We observe that regardless of the model employed in the second stage, the HM@K metric consistently outperforms the baseline M@K across all tested benchmarks (AIME25 and HMMT25-Feb).
This empirical evidence suggests that the heavy thinking framework is highly versatile and compatible with diverse model architectures, effectively enhancing reasoning performance through cross-model collaboration.

Furthermore, we highlight a counter-intuitive finding regarding the model choice for the second stage. 
While Qwen2.5-32B-Instruct itself does not exhibit superior performance in solving these complex reasoning problems independently~\citep{Yang2025Qwen3}~\footnote{From the official result in Qwen3 technique report, Qwen2.5-32B-Instruct achieves 12.8\% accuracy on AIME25, which is lower than R1-Distill-Qwen-7.}, its integration into the sequential deliberation phase yields results that align with our expectations of performance gains. 
This observation leads to a crucial insight: the sequential deliberation phase does not necessarily demand peak intrinsic reasoning power from the model. 
Instead, it relies more heavily on the model's ability to perform comprehensive analysis, synthesis, and summarization of the thought traces generated in the first stage. 
This suggests that the heavy thinking paradigm can be effectively scaled by utilizing larger, more instruction-following models for deliberation, even if their specialized reasoning capabilities are not the primary driver of success.

\subsection{Effectiveness of Iteration Deliberation}

In our heavy thinking framework, we introduce an iterative deliberation mechanism to allow the model to recursively analyze parallel trajectories by incorporating previously generated summary information. 
In our experiments, we fix the number of outputs for both parallel reasoning and sequential deliberation at $K=K^{(1)}=\cdots=K^{(N)}=8$, and $N$ is set as 4.
To evaluate the robustness and scalability of this approach, we conduct experiments across three models of varying scales, i.e., R1-Distill-Qwen-7B, R1-Distill-Qwen3-8B, and DeepSeek-R1-0528.

As illustrate in Figure~\ref{fig:iterative_heavy}, we observe a consistent upward trend in the HM@K metric as the number of iterations increases. 
This phenomenon suggests that the heavy thinking framework exhibits intrinsic scaling capabilities, where extended deliberation cycles contribute to improved collective reasoning performance.
However, this gain is accompanied by a significant degradation in the HP@K metric. 
This divergence indicates that while iterative processing helps in certain dimensions, subsequent deliberation steps may be susceptible to interference from information generated in earlier stages. 
Such interference likely introduces cumulative noise or biases that constrain the model's refinement space, ultimately limiting the potential for further performance gains.
These findings highlight a critical trade-off between iterative depth and information consistency within the deliberation process.

\subsection{Adaptation to Agentic Tool Use}

To further explore the generalizability of the heavy thinking framework, we investigate its performance in scenarios requiring external tool interactions. Specifically, we generate reasoning trajectories that incorporate tool calls during the parallel reasoning phase. For this experiment, we select three models equipped with native tool-calling capabilities: Qwen3-8B, Qwen3-32B, and GPT-OSS-20B. We utilize a Python interpreter to provide execution feedback, which serves as a crucial signal for the deliberation process. The maximum number of interaction rounds between the model and the interpreter is capped at 50 to ensure a balance between reasoning depth and computational efficiency.

Table~\ref{tab:heavy-tir} summarizes the comparative results on the AIME25 and HMMT25 benchmarks. The empirical evidence demonstrates that the heavy thinking mode (represented by HM@4) consistently surpasses the performance of the traditional majority voting baseline (V@4) across all tested models and datasets. 
For instance, on the AIME25 benchmark, GPT-OSS-20B achieves an accuracy of 90.0\% under the heavy thinking framework, significantly exceeding the 83.3\% achieved by voting. 
These results indicate that the sequential deliberation mechanism effectively leverages the feedback signals from tool execution, allowing the model to refine its trajectories more accurately.

\section{Related Work}

\subsection{Parallel Reasoning in LLM}
Parallel thinking has recently emerged as a vibrant research frontier, which can be viewed as a efficient but effect test-time scaling technique~\citep{Pan2025Learning, Liu2024APAR, Jin2025Learning, Rodionov2025Hogwild, Yang2025Multiverse, Yang2025APE}. 
Predominant among current approaches are brute-force strategies that either spawn independent trajectories with end-stage aggregation~\citep{Brown2024Large, Zheng2025Parallel, Wen2025ParaThinker} or synchronize thoughts at rigid, pre-defined intervals~\citep{Hsu2025Group, Macfarlane2025Instilling}. 
Such paradigms inherently lack adaptivity, as their branching and merging points are dictated by static schedules rather than the evolving intermediate progress of the reasoning process. 
While frameworks like Monte Carlo Tree Search~\citep{Zhang2024Accessing} and Tree of Thoughts~\citep{Yao2023Tree} provide more granular parallelization, they remain tethered to hand-crafted heuristics and external verifiers. 
In contrast, the recently emerging heavy thinking approach offers more flexibility in implementing test time scaling, as demonstrated in Gemini~\citep{deepmind2024geminiimo} and Kimi K2~\citep{Bai2025Kimi}, and PaCoRe~\citep{pacore2025StepFun}. This paper delves into the specific implementation of the heavy thinking pattern and analyzes its effectiveness across multiple tasks in various domains.

\begin{table}[!t]
\footnotesize
\renewcommand{\arraystretch}{1.5}

\setlength{\tabcolsep}{0.8mm}
\resizebox{\linewidth}{!}{
\begin{tabular}{c | cccc | cccc }
\toprule
\multirow{2}{*}{\textbf{Models}}
& \multicolumn{4}{c|}{\textbf{AIME25}} 
& \multicolumn{4}{c}{\textbf{HMMT25}} \\

\cmidrule(lr){2-5} \cmidrule(lr){6-9}
& \makecell{M$@k$}
& \makecell{P$@k$}
& \makecell{V$@4$}
& \makecell{HM$@4$}
& \makecell{M$@k$}
& \makecell{P$@k$}
& \makecell{V$@4$}
& \makecell{HM$@4$}
\\
\midrule

\makecell{Qwen3-8B} &
55.7 & 83.3 & 68.3 & 76.7 &
 38.0 & 70.7 & 54.1 & 69.3 \\

\midrule

\makecell{Qwen3-32B} &
63.0 & 89.0 & 83.3 & 80.0 &
40.3 & 81.7 & 63.3 & 68.5 \\

\midrule

\makecell{GPT-\\OSS-20B} &
69.8 & 95.0 & 83.3 & 90.0 &
55.3 & 93.3 & 73.3 & 85.7 \\
 
\bottomrule
\end{tabular}
}
\centering
\caption{Overview performance of heavy mode on tool-interleave reasoning scenarios (Heavy Mean@4, simplify as HM@4) compared to basic TTS metrics, i.e., Mean@K (M@K),  Pass@K (P@K), and Vote@K (V@K).}
\label{tab:heavy-tir} 
\end{table}

\subsection{Test-time Scaling in LLM Post-training}
With the development of OpenAI's o1~\citep{Jaech2024OpenAI}, DeepSeek R1~\citep{guo2025deepseek}, and Gemini~\citep{deepmind2024geminiimo}, the post-training with test-time scaling has been powerful and versatile techniques in LLM reasoning.
Theses approaches aims to develop the LLM with capabilities for correcting, reflecting, critiquing, and verifying by themselves, typically increasing inference computation by extending the model’s reasoning chains through long CoT~\citep{Wei2022Chain, Zhang2024ReST, Wang2024Self, Zelikman2024Quiet, Lightman2024Let, Wang2025Prejudge, Wang2025Boosting}.
Recently, multiple efforts devote to improve the reasoning capabilities by introducing efficiency and stability RLVR, which aims to optimize LLM via outcome-based, automatically checkable rewards, or step-wise signals~\citep{Yu2025DAPO, Yue2025VAPO, Liu2025SPIRAL, Chen2025MiniMax}.
In this paper, we introduce RLVR in heavy thinking to investigate whether the LLM can break the reasoning boundaries.

\section{Conclusion}

In this paper, we have systematically investigated heavy thinking as a novel TTS strategy to enhance the reasoning capabilities of LLMs, and distill the whole workflow into  a readable skill for agentic harness.
By introducing a framework centered on parallel reasoning and sequential deliberation, we provided a clear structural understanding of how test-time computation translates into superior task performance. 
Our extensive evaluations reveal that heavy thinking consistently outperforms traditional Best-of-N strategies (e.g., voting method), particularly in models with high intrinsic reasoning potential where performance can approach the theoretical Pass@K limit. 
Extensive detailed analysis is also conducted to show the effect of the heavy thinking. Most importantly, our findings demonstrate that RLVR can also substantially improve the model's reasoning capabilities.
In the future, we aim to conduct a more granular analysis of the performance of heavy thinking trajectories within RL frameworks.

\bibliography{example_paper}
\bibliographystyle{icml2026}

\newpage

\appendix

\section{Impact of Parallel Trajectories}

\begin{figure}[t]
\begin{center}
  \includegraphics[width=.95\linewidth]{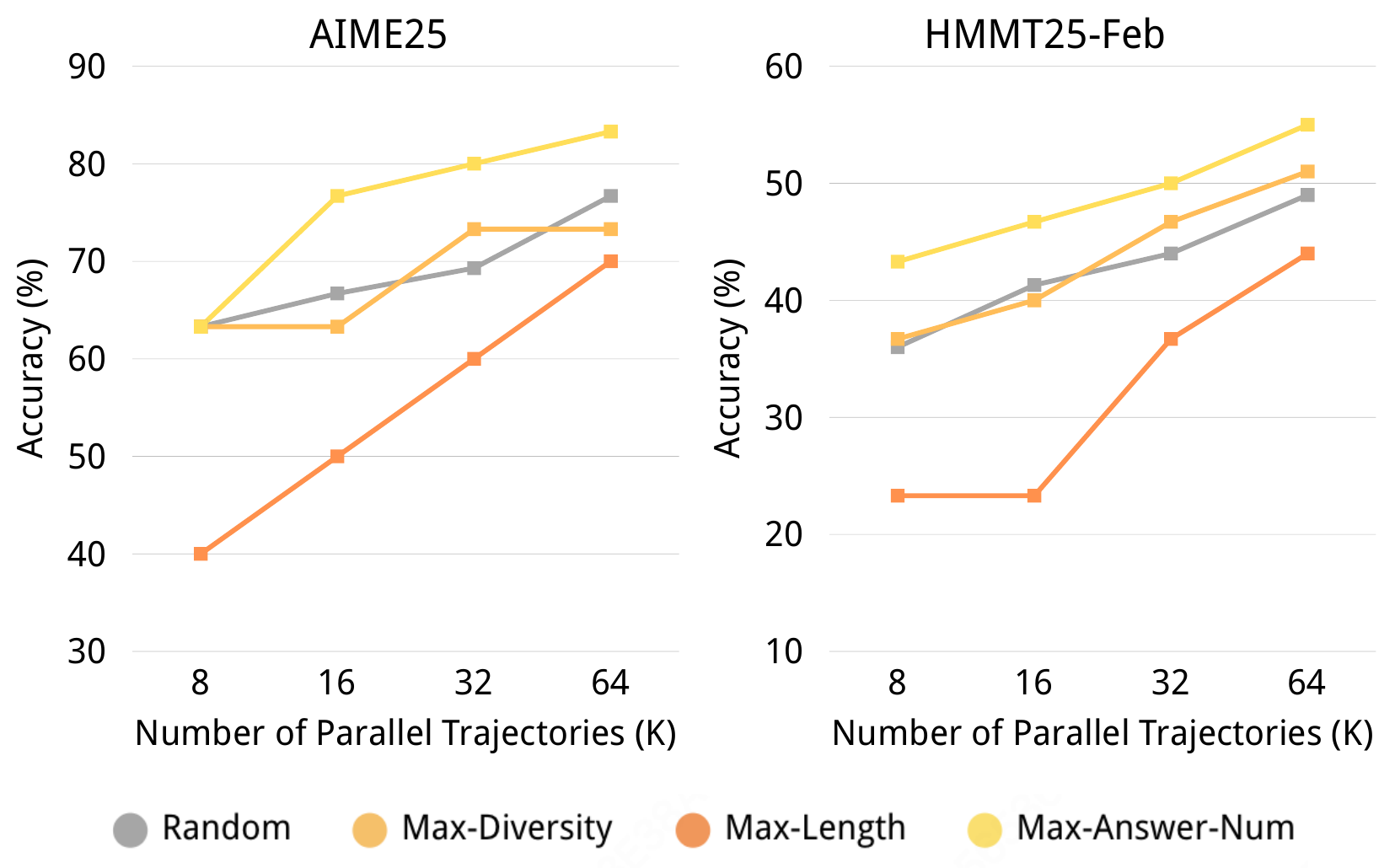} \hfill
  \caption{When choosing different permutations. Random: randomly select $K$ trajectories; Max-Diversity: select $K$ trajectories that have the highest diversity; Max-Length: select the top $K$ trajectories based on the length; Max-Answer-Num: select the trajectories that have the highest frequency answer.}
  \label{fig:heavy_permutations}
\end{center}
\end{figure}

In this section, we aim to investigate the impact of the permutations of parallel reasoning. 
We select R1-Distill-Qwen3-8B model as the basic backbone, and evaluate across two challenging benchmarks, including AIME25 and HMMT25-Feb. 
For each problem, we generate 256 parallel trajectories, and compare four distinct trajectory selection (permutation) strategies: Random, Max-Diversity, Max-Length, and Max-Answer-Num.

The experimental results are illustrated in Figure~\ref{fig:heavy_permutations}, we draw the following suggestions. 1) Consistent upward trend in accuracy for all selection methods as the number of parallel trajectories K increases. 
This demonstrates that increasing the inference budget through parallelization consistently benefits model performance, regardless of the selection heuristic employed.
2) Interestingly, the Max-Diversity strategy yields performance comparable to Random sampling. This suggests that while diversity is a natural byproduct of increased sampling temperatures, explicitly optimizing for trajectory diversity does not provide a significant marginal gain. Conversely, the Max-Length strategy performs the worst among all tested methods. This indicates that a preference for longer outputs introduces substantial noise, suggesting that verbosity does not necessarily correlate with reasoning quality in parallel inference settings.
3) The Max-Answer-Num strategy significantly outperforms all other baselines. 
This result highlights that leveraging consensus (majority voting) to select parallel trajectories effectively increases the proportion of correct reasoning paths within the candidate set. Consequently, this provides a more robust foundation for the heavy thinking module to perform synthesis, summarization, and final logical deduction.

\begin{figure*}[t]
  \includegraphics[width=\linewidth]{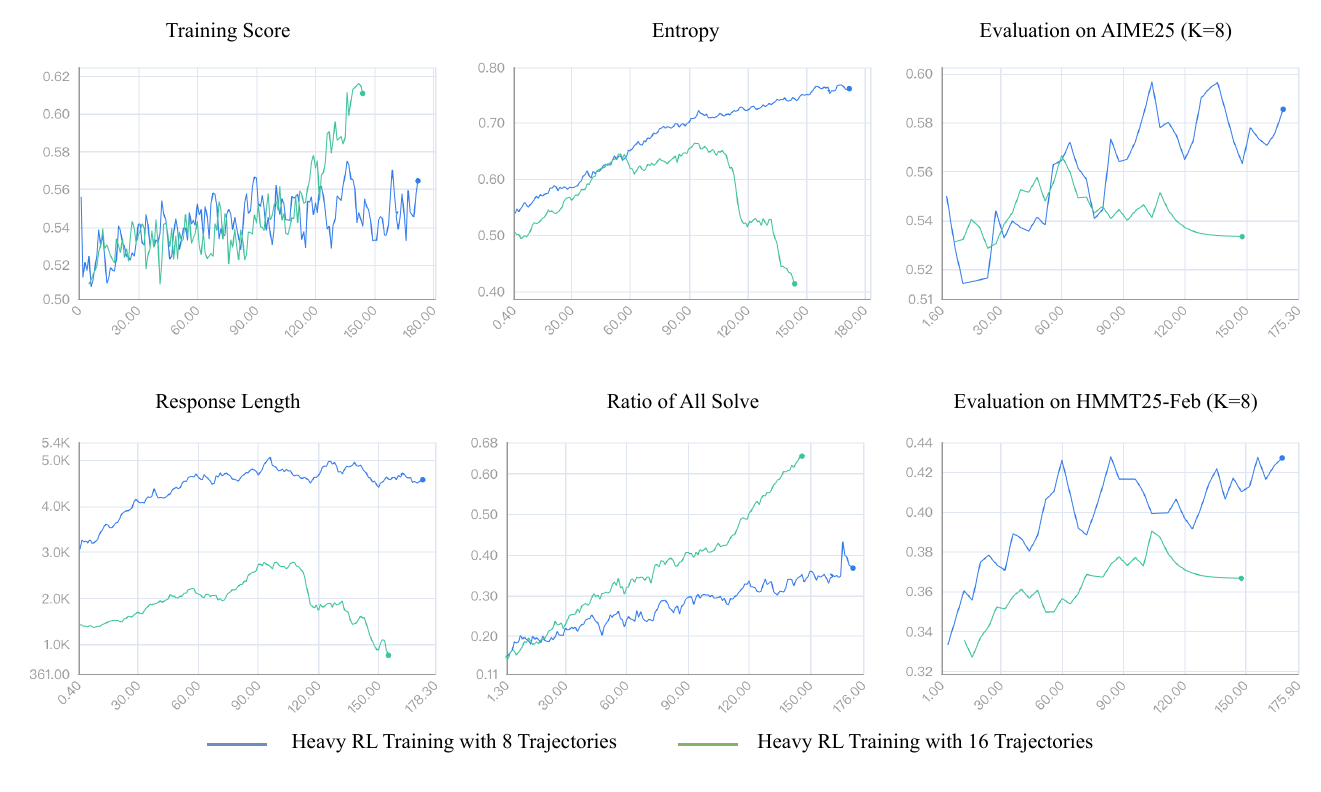} \hfill
  \caption{The RL training on heavy thinking. Blue curve denotes the number of parallel trajectories is 8, while the green curve denotes 16. We use the VeRL framework to perform RLVR.}
  \label{fig:heavy_rl}
\end{figure*}

\section{Advancing Heavy Thinking via RLVR}
Through extensive preliminary experiments, we observe that under the "heavy thinking" mode, the HP@K metric can significantly outperform HM@K in specific scenarios. This observation motivates a critical research question: can applying Reinforcement Learning from Verifiable Rewards (RLVR) directly to heavy thinking trajectories further elevate the upper bound of the model's reasoning capabilities?

To investigate this, we reuse the parallel reasoning trajectories generated in Experiment 3. We specifically select queries with a pass rate in the range $[0,0.625]$ and randomly sample $K\in\{8,16\}$ trajectories to be encapsulated as serialized memory caches. Our experimental framework is built upon VeRL~\citep{Sheng2025HybridFlow}, employing GSPO~\citep{Zheng2025Group} as the reinforcement learning algorithm. We utilize R1-Distill-Qwen-7B as the backbone model. 

The experimental results are illustrated in Figure~\ref{fig:heavy_rl}. During the initial training phase (within the first 100 steps), the model exhibits a consistent growth trend on both the training and test sets. Notably, the HM@4 metric achieves a further improvement of approximately 10\%.
However, we observe a distinct divergence in stability between the two configurations. For the $K=16$ group, the model experiences a significant entropy collapse after 100 steps. In contrast, the $K=8$ configuration demonstrates superior stability throughout the training process. We think that this instability in the $K=16$ setting is primarily attributed to the maximum sequence length limitations of the R1-Distill-Qwen-7B model, which may lead to truncated or suboptimal training signals when handling longer serialized contexts.
While the initial results are promising, the dynamics of heavy thinking under RLVR require further exploration.

\section{Prompt and SKill for Heavy Thinking}
\label{app:memory}

\begin{figure*}[t]
  \includegraphics[width=\linewidth]{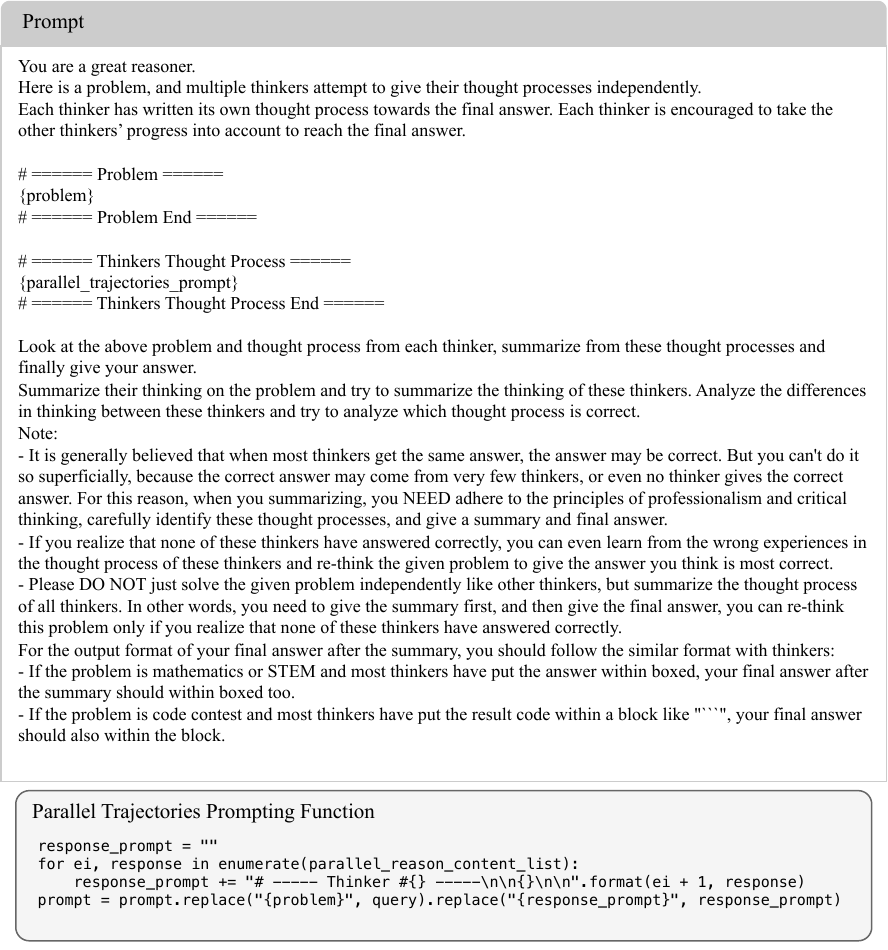} \hfill
  \caption{The prompt of heavy thinking in serialized memory cache.}
  \label{fig:heavy_prompt}
\end{figure*}

The prompt for memory cache is shown in Figure~\ref{fig:heavy_prompt}.

\begin{figure*}[t]
  \includegraphics[width=\linewidth]{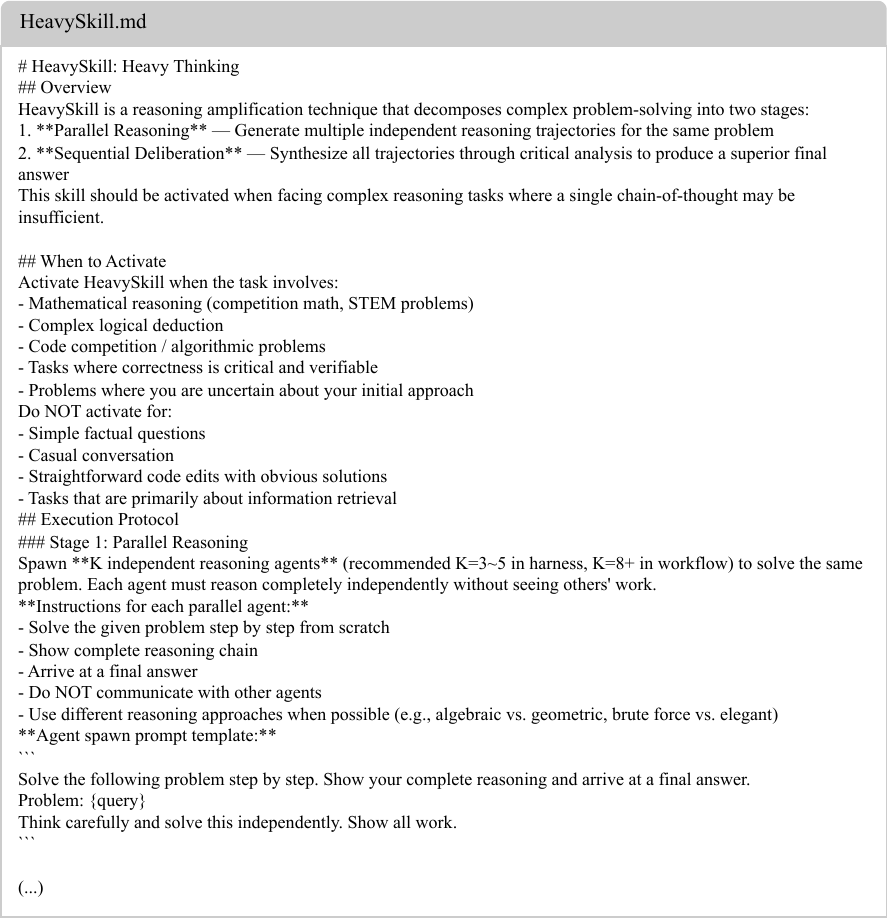} \hfill
  \caption{The skill file of heavy thinking (part I).}
  \label{fig:heavy_skill_1}
\end{figure*}

\begin{figure*}[t]
  \includegraphics[width=\linewidth]{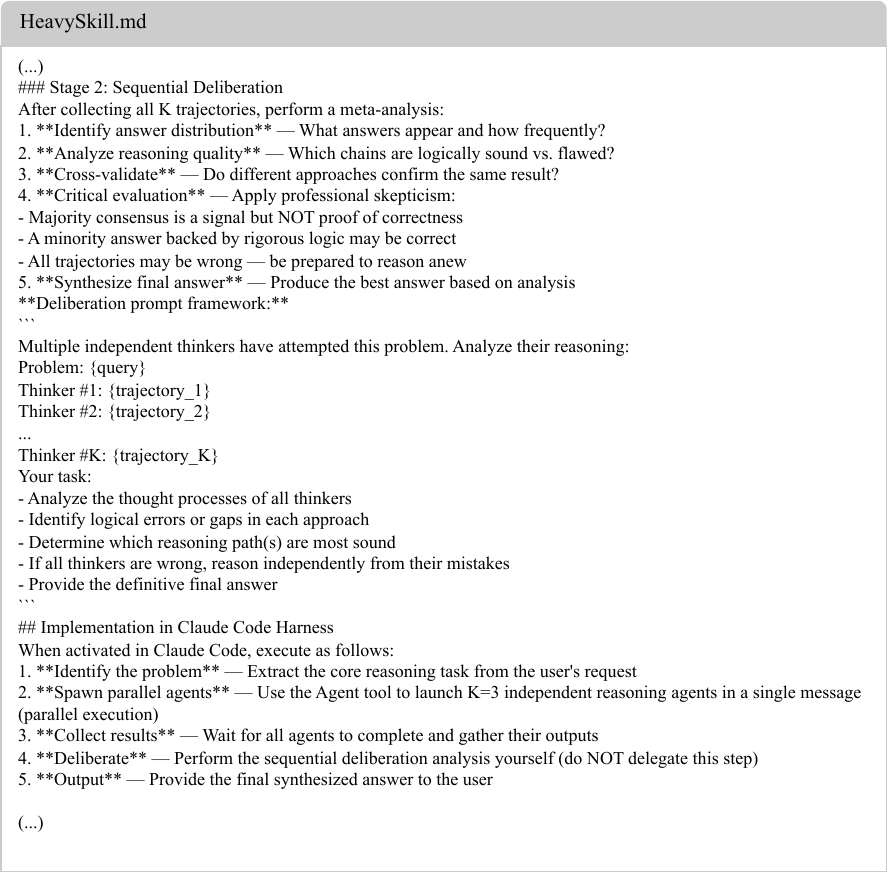} \hfill
  \caption{The skill file of heavy thinking (part II).}
  \label{fig:heavy_skill_2}
\end{figure*}

\begin{figure*}[t]
  \includegraphics[width=\linewidth]{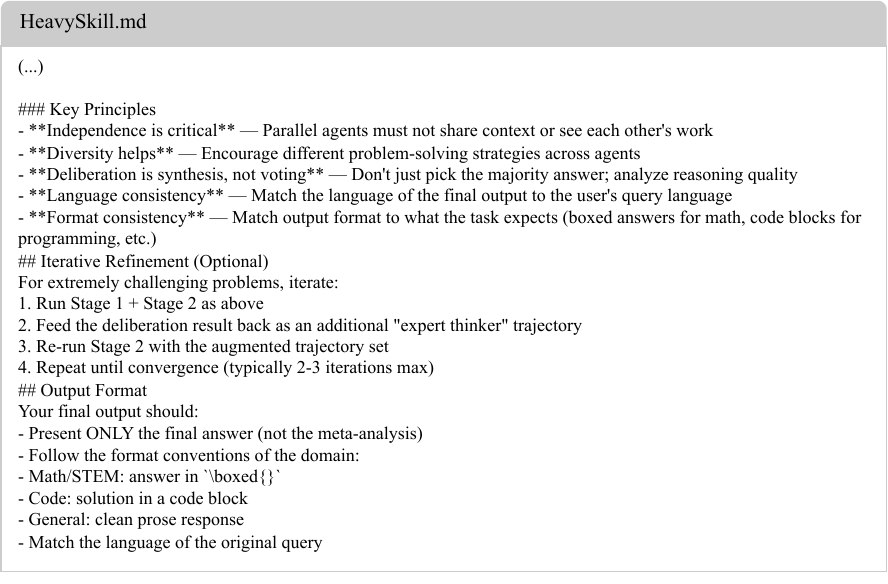} \hfill
  \caption{The skill file of heavy thinking (part III).}
  \label{fig:heavy_skill_3}
\end{figure*}

The skill file for agentic harness is shown in Figure~\ref{fig:heavy_skill_1}, Figure~\ref{fig:heavy_skill_2}, and Figure~\ref{fig:heavy_skill_3}.

\end{document}